\documentclass[10pt,twocolumn,letterpaper]{article}

\usepackage{cvpr}
\usepackage{times}
\usepackage{epsfig}
\usepackage{graphicx}
\usepackage{amsmath}
\usepackage{amssymb}
\usepackage{graphics}
\usepackage{booktabs}
\usepackage{multirow}
\usepackage{enumitem}
\usepackage{bm}
\usepackage{adjustbox}
\usepackage[innercaption]{sidecap}
\usepackage{upquote}
\usepackage{textcomp}
\usepackage{pifont}
\usepackage{etoolbox}
\usepackage{caption}
\usepackage{subcaption}
\usepackage[utf8]{inputenc}
\usepackage{xspace}
\usepackage{enumitem}

\hypersetup{
	pagebackref=true,
	breaklinks=true,
	letterpaper=true,
	colorlinks=true,
	bookmarks=false,
}

\sidecaptionvpos{figure}{m}

\graphicspath{{./}}

\DeclareGraphicsExtensions{.jpeg,.png,.pdf}

\newcommand{\cmark}{\ding{51}}%
\newcommand{\timesnarrow}{{\mkern-2mu\times\mkern-2mu}}

\newcommand{\partitle}[1]{\noindent\textbf{#1}}
\newcommand{\ptspace}{\vspace*{5pt}}

\DeclareFontFamily{U}{mathb}{\hyphenchar\font45}
\DeclareFontShape{U}{mathb}{m}{n}{
<-6> mathb5
<6-7> mathb6
<7-8> mathb7
<8-9> mathb8
<9-10> mathb9
<10-12> mathb10
<12-> mathb12}{}
\DeclareSymbolFont{mathb}{U}{mathb}{m}{n}
\DeclareMathSymbol{\llcurly}{\mathrel}{mathb}{"CE}
\DeclareMathSymbol{\ggcurly}{\mathrel}{mathb}{"CF}

\cvprfinalcopy

\def\cvprPaperID{}

\ifcvprfinal\fi
\begin{document}

\title{Timeception for Complex Action Recognition}

\author{
Noureldien~Hussein, Efstratios~Gavves, Arnold~W.M.~Smeulders
\\
QUVA~Lab, University~of~Amsterdam
\\
{\tt\small\{nhussein,egavves,a.w.m.smeulders\}@uva.nl}}

\maketitle

\begin{abstract}
This paper focuses on the temporal aspect for recognizing human activities in videos; an important visual cue that has long been undervalued.
We revisit the conventional definition of activity and restrict it to ``Complex Action": a set of one-actions with a weak temporal pattern that serves a specific purpose.
Related works use spatiotemporal 3D convolutions with fixed kernel size, too rigid to capture the varieties in temporal extents of complex actions, and too short for long-range temporal modeling.
In contrast, we use multi-scale temporal convolutions, and we reduce the complexity of 3D convolutions. The outcome is Timeception convolution layers, which reasons about minute-long temporal patterns, a factor of 8 longer than best related works.
As a result, Timeception achieves impressive accuracy in recognizing the human activities of Charades, Breakfast Actions, and MultiTHUMOS.
Further, we demonstrate that Timeception learns long-range temporal dependencies and tolerate temporal extents of complex actions.
\vspace{-15pt}
\end{abstract}

\section{Introduction}
\begin{figure}[!ht]
\begin{center}
\includegraphics[trim=0mm 6mm 0mm 0mm,width=0.85\linewidth]{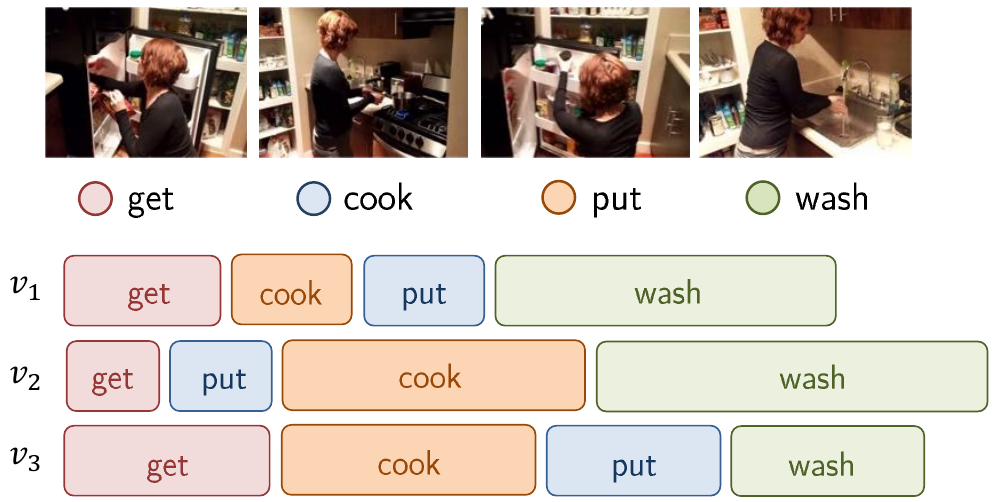}
\end{center}
\caption{Properties of complex action ``Cooking a Meal":\\
\textbf{composition}: consists of several one-actions (Cook, ...),
\protect\linebreak
\textbf{order}: weak temporal order of one-actions (Get {\large $\llcurly$} Wash),
\protect\linebreak
\textbf{extent}: one-actions vary in their temporal extents.}
\label{fig:1-1}
\vspace*{-5mm}
\end{figure}

In ordinary life, activities of daily living pop up frequently.
Our conversations include actions like ``cooking a meal'' or ``cleaning the house'' much more frequently than actions like ``jumping'' or ``cutting a cucumber''.
The latter, which we call \emph{one-actions}, exhibit one visual pattern, possibly repetitive.
They are usually short in time, homogeneous in motion and coherent in form.
In contrast, cooking a meal or cleaning the house are very different actions.
We refer to them as \emph{complex actions}, characterized by:
\emph{i.} They are typically composed of several one-actions, see  figure~\ref{fig:1-1}.
\emph{ii.}
These one-actions, contained in a complex action, exhibit large variations in their temporal duration and temporal order.
\emph{iii.} As a consequence of the composition, a complex action takes much longer to unfold.
And, by the in-homogeneity in composition, the complex action needs to be sampled in full, not to miss crucial parts.

In the recent literature, the main focus is the recognition of short-range actions like in HMDB, UCF and Kinetics \cite{kuehne2011hmdb,soomro2012ucf101,kay2017kinetics}. Few attention has been paid to the recognition of long-range and complex actions, as in Charades and EventNet \cite{sigurdsson2016hollywood,ye2015eventnet}, which we study here. The first challenge is minute-long temporal modeling while maintaining attention to seconds-long details. Statistical temporal pooling, as applied in  \cite{girdhar2017attentional,miech2017learnable} falls short of learning temporal order. Neural temporal modeling~\cite{donahue2015long,girdhar2017actionvlad} and spatio-temporal convolutions of various types~\cite{ji20133d,tran2015learning, tran2018closer} successfully learns temporal order of 8~\cite{carreira2017quo} or 128 timesteps~\cite{wang2017non}. But the computational cost is far beyond scaling up to 1000 timesteps needed for complex actions. The second challenge is tolerating variations in temporal extent and temporal order of one-actions. Related methods~\cite{tran2015learning,xie2017rethinking} learn spatio-temporal convolutions with fixed-size kernels, which would be too rigid for complex actions. To address these challenges, we present Timeception, a novel convolutional layer dedicated only for temporal modeling. It learns long-range temporal dependencies with attention to short-range details. Plus, it tolerates the differences in temporal extent of one-actions comprising the complex action. As a result, we demonstrate success in recognizing the long and complex actions, and achieving state-of-the-art-results in Charades~\cite{sigurdsson2016hollywood}, Breakfast Actions~\cite{kuehne2014language} and MultiTHUMOS~\cite{yeung2018every}.

The novelties of of this paper are:
\textit{i.} We introduce a convolutional temporal layer effectively and efficiently learn minute-long action ranges of 1024 timesteps, a factor of 8 longer than best related work.
\textit{ii}. We introduce multi-scale temporal kernels to account for large variations in duration of action components.
\textit{iii}. We use temporal-only convolutions, which are better suited for complex actions than spatiotemporal counterparts.

\begin{figure*}[!ht]
\begin{center}
\includegraphics[trim=0mm 8mm 0mm 5mm,width=0.7\linewidth]{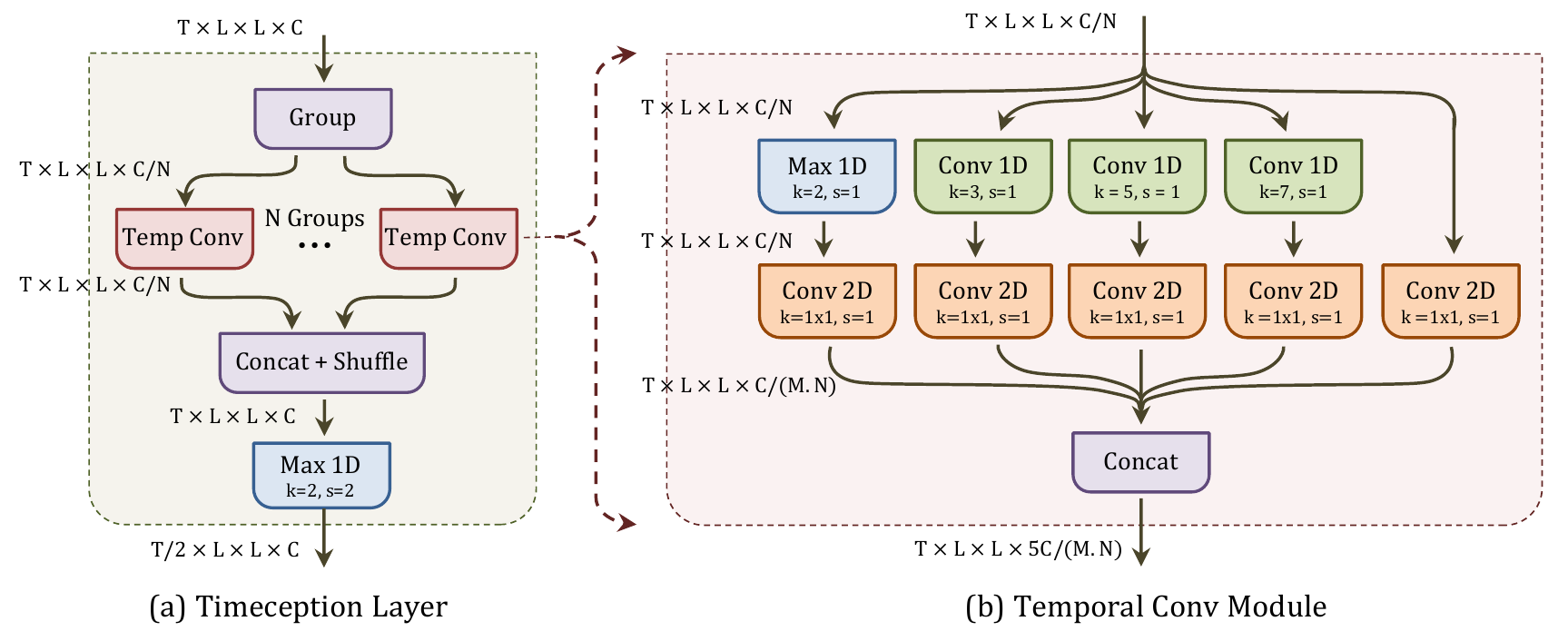}
\end{center}
\caption{The core component of our method is Timeception layer, left.
Simply, it takes as an input the features $\mathbf{X}$; corresponding to $T$ timesteps from the previous layer in the network.
Then, it splits them into $N$ groups, and temporally convolves each group using temporal convolution module, right.
It is a novel building block comprising multi-scale \textit{temporal-only} convolutions to tolerate a variety of temporal extents in a complex action.
Timeception makes use of grouped convolutions and channel shuffling to learn cross-channel correlations efficiently than $1 \timesnarrow 1$ spatial convolutions.}
\label{fig:3-1}
\vspace*{-5mm}
\end{figure*}

\section{Related Work}\label{sec:related_work}
\partitle{Temporal Modeling.}
The stark difference between video and image classification is the temporal dimension, which necessitates temporal modeling.
A widely used approach is statistical pooling: max and average pooling~\cite{hussein2017unified,habibian2017video2vec}, attention pooling~\cite{girdhar2017attentional}, rank pooling~\cite{fernando2017rank},  dynamic images~\cite{bilen2016dynamic} and context gating~\cite{miech2017learnable}, to name a few.
Beyond statistical pooling, vector aggregation is also used.
~\cite{oneata2013action} uses Fisher Vector~\cite{sanchez2013image} to aggregate spatio-temporal features over time,
while~\cite{girdhar2017actionvlad,cosmin2017spatio,duta2017spatio} extend VLAD~\cite{arandjelovic2013all} to use local convolution features extracted from video frames.
The downside of statistical pooling and vector aggregation is completely neglecting temporal patterns -- an important visual cue.

Other strands of work use neural methods for temporal modeling.
LSTMs are used to model the sequence in action videos~\cite{donahue2015long}.
While TA-DenseNet\cite{ghodrati2018video} extents DenseNet~\cite{huang2017densely} to exploit the temporal dimension.
To our knownedge, no substantial improvements have been reported recently.

\partitle{Short-range Action Recognition.}
Few works~\cite{karpathy2014large} learn deep appearance features by frame-level classification of actions, using 2D CNNs.
Others complement deep appearance features with shallow motion features, as IDT~\cite{feichtenhofer2016convolutional}.
Also, auxiliary image representations is fused with RGB signals:
~\cite{simonyan2014two} uses OpticalFlow channels, while~\cite{bilen2017action} use Dynamic Images.
3D CNNs are the natural evolution of their 2D counterparts.
C3D~\cite{tran2015learning,ji20133d} proposes 3D CNNs to capture spatio-temporal patterns of 8 frames in a sequence.
In the same vein, I3D ~\cite{carreira2017quo} inflates the kernels of ImageNet-pretrained 2D CNN to jump-start the training of 3D CNNs.
While effective in short-range video sequences of few seconds, 3D convolutions are too computationally expensive to address minute-long videos, which is our focus.

\partitle{Long-range Action Recognition.}
To learn long-range temporal patterns, ~\cite{sigurdsson2017asynchronous} uses CRF on top of CNN feature maps to model human activities.
To learn video-wide representations, TRN~\cite{zhou2017temporal} learns relations between several video segments.
TSN~\cite{wang2016temporal,wang2018temporal} learns temporal structure in long videos.
LTC~\cite{varol2017long} considers different temporal resolutions as a substitute to bigger temporal windows.
Inspired by self-attention~\cite{vaswani2017attention}, non-local networks~\cite{wang2017non} proposes a 3D CNN with a long temporal footprint of 128 timesteps.

All aforementioned methods succeed in modeling temporal footprint of 128 timesteps ($\sim$4-5 sec) at max.
In this work, we address complex actions with long-range temporal dependencies of up to 1024 timesteps, jointly.

\partitle{Convolution Decomposition.}
CNNs succeed in learning spatial~\cite{wang2015towards,karpathy2014large} and spatiotemporal~\cite{,tran2015learning,ji20133d,feichtenhofer2018have,varol2017long,hara2017can} action concepts,
but existing convolutions grow heavy in computation, specially at the higher layers where the number of channels can grow as much as 2k~\cite{he2016deep}.
To control the computational complexity, several works propose the decomposition of 2D and 3D convolutions.
Xception~\cite{chollet2016xception} argues that separable 2D convolutions are as effective as typical 2D convolutions.
Similarly, S3D~\cite{xie2017rethinking, tran2018closer} considers separable 2+1D convolutions to reduce the complexity of typical 3D convolutions.
ResNet~\cite{he2016deep} reduces the channel dimension using $1 \timesnarrow 1$ 2D convolution before applying the costly $3 \timesnarrow 3$ 2D spatial convolution.
ShuffleNet~\cite{zhang2017shufflenet} models cross-channel correlation by channel shuffling instead of $1 \timesnarrow 1$ 2D convolution.
ResNeXt~\cite{xie2017aggregated} proposes grouped convolutions, while Inception~\cite{szegedy2015going,szegedy2016rethinking} replaces the fixed-size 2D spatial kernels into multi-scale 2D spatial kernels of different sizes.

In this work, we propose the decomposition of spatiotemporal convolutions into depthwise-separable temporal convolutions, which we show to be better suited for long-range temporal modeling that 2+1D convolutions.
Moreover, to account for the differences in temporal extents, we propose temporal convolutions with multi-scale kernels.

\section{Method}\label{sec:method}
\subsection{Motivation}
\label{subsec:motivation}
Modern 3D CNNs learn spatiotemporal kernels over three orthogonal subspaces of video information: the temporal ($\mathcal{T}$), the spatial ($\mathcal{S}$) and the semantic channel subspace ($\mathcal{C}$).
One spatiotemporal kernel $w \in \mathbb{R}^{T \times L \times L \times C}$ learns a latent concept by simultaneously convolving these three subspaces~\cite{tran2015learning, carreira2017quo}, where $T$ is the number of timesteps, $C$ is the number of channels, and $L$ is the size of spatial window.
Though, there is no fundamental reason why these subspaces must be convolved simultaneously.
Instead, as showcased in~\cite{xie2017rethinking}, one can model these subspaces separately, $w \propto w_{s} \times w_{t}$, by decomposing $w$ into spatial $w_{s} \in \mathbb{R}^{1 \times L \times L \times C}$ and temporal $w_{t} \in \mathbb{R}^{T \times 1 \times 1 \times C}$ kernels.
Strictly speaking, while replacing $w$ with a cascade $\tilde{w}=w_s \times w_t$ is often referred to as \emph{``decomposition''}, this operation is not tensor decomposition -- there is no strict requirement that, at optimality, we have $w^* \equiv \tilde{w}^*$.
Instead, as the cascade $\tilde{w}$ is, by definition, computationally more efficient than the full kernel $w^{}$, the only practical requirement is that the resulting cascade $\tilde{w}$ yields equally good or better accuracies for the task at hand.
In light of this realization, while the aforementioned decomposition along the spatial and temporal axes is intuitive and empirically successful~\cite{xie2017rethinking}, it is not the only possibility.
Therefore, Any other decomposition is permissible, namely:
$\tilde{w}= w_\alpha \times w_\beta \times w_\gamma \times ...$,
as long as some basic principles are maintained for the final cascade $\tilde{w}$.
Generalizing on recent decomposed architectures ~\cite{tran2018closer,chollet2016xception}, we identify from the literature three intuitive design principles for the spatiotemporal CNNs:

\partitle{\textit{i}. Subspace Modularity.}
In the context of deep network cascades, a decomposition should be modular, such that between subspaces, it retains the nature of the respective subspaces across subsequent layers.
Namely, after a cascade of spatial and a temporal convolutions, it must be possible that yet another cascade (of spatial and temporal convolutions) is possible and meaningful.

\partitle{\textit{ii}. Subspace Balance.}
A decomposition should make sure that a balance is retained between the subspaces and their parameterization in different layers.
Namely, increasing the number of parameters for modeling a specific subspace should come at the expense of reducing the number of parameters of another subspace. 
A typical example is conventional 2D CNN, in which the spatial subspace ($\mathcal{S}$) is reduced while the semantic channel subspace ($\mathcal{C}$) is expanded.

\partitle{\textit{iii}. Subspace Efficiency.}
When designing the decomposition for a specific task, we should make sure that the bulk of the available parameter budget is dedicated to subspaces that are directly relevant to the task at hand.
For instance, for long-range temporal modeling, a logical choice is a decomposition that increases the convolutional parameters for the temporal subspace ($\mathcal{T}$).

Motivated by the aforementioned design principles, we propose a new temporal convolution layer for encoding long-range patterns in complex actions, named Timeception, see figure~\ref{fig:3-1}.
First, we discuss the Timeception layer.
Then we describe how to stack Timeception layers on top of existing 2D or 3D CNNs.

\subsection{Timeception Layer}
For modeling complex actions in long videos, our temporal modeling layer faces two objectives.
First, we would like to learn the possible \emph{long-range temporal dependencies} between one-actions throughout the entire video, and for a frame sequence of up to 1000 timesteps.
Second, we would like to \emph{tolerate the variations in the temporal extents} of one-actions throughout the video.

Next, we present the Timeception layer, designed with these two objectives in mind.
Timeception is a layer that sits on top of either previous Timeception layers, or a CNN.
The CNN can be either purely spatial; processing frames independently, like ResNet~\cite{he2016deep}, or short-range spatiotemporal; processing nearby bursts of frames, like I3D~\cite{carreira2017quo}.

\partitle{Long-range Temporal Dependencies.}
There exist two design consequences for modeling long-range temporal dependencies between one-actions throughout the video. The first consequence is that our temporal network must be composed of deeper stacks of temporal layers.
Via successive layers, thereafter, complex and abstract spatiotemporal patterns can emerge, even when they reside at temporally very distant locations in the video.
Given that we need deeper temporal stacks and we have a specific parameter budget for the complete model, the second consequence is that the temporal layers must be as cost-effective as possible.

Revisiting the cost-effectiveness of spatiotemporal models, existing architectures rely either on joint spatiotemporal kernels~\cite{carreira2017quo} with parameter complexity $\mathcal{O}(T\cdot L^2\cdot C)$ or decomposed spatial and temporal kernels~\cite{xie2017rethinking,tran2018closer} with parameter complexity $\mathcal{O}((L^2+T)\cdot C)$.
To make the Timeception layer temporally cost-effective, according to the third design principle of \emph{subspace importance}, we opt for trading spatial and semantic complexity for longer temporal windows.
Specifically, we propose depthwise-separable temporal convolution with kernel $w^{TC}_{t} \in \mathbb{R}^{T \timesnarrow 1 \timesnarrow 1 \timesnarrow 1}$. Hereafter, we refer to this convolution as \emph{temporal-only}.
What is more, unlike~\cite{carreira2017quo,xie2017rethinking,tran2018closer}, we propose to focus only on temporal modeling and drop the spatial kernel $w_{s} \in \mathbb{R}^{1 \timesnarrow L \timesnarrow L \timesnarrow C}$ altogether.
Hence, the Timeception layer relies completely on the preceding CNN for the detection of any spatial pattern.

The simplified \emph{temporal-only} kernel has some interesting properties.
Each kernel acts on only one channel.
As the kernels do not extend to the channel subspace, they are encouraged to learn generic and abstract, rather than semantically-specific, temporal combinations.
For instance, the kernels learn to detect the temporal pattern of one \emph{latent concept} represented by one channel.
Last, as the parameter complexity of a single Timeception layer is approximately $\mathcal{O}(T + log L)$, it is computationally feasible to train a deep model to encode temporal patterns of up to 1024 timesteps. 
This amounts to about 40 seconds of video sequences.

Unfortunately, by stacking \emph{temporal-only} convolutions one after the other, we violate the first design principle of \emph{subspace modularity}.
The reason is that the semantic subspace in long-range spatiotemporal patterns is ignored.
To this end, we propose to use \emph{channel grouping} operation~\cite{xie2017aggregated} before the temporal-only convolutions and \emph{channel shuffling} operation~\cite{zhang2017shufflenet} after the temporal-only convolutions.
The purpose of channel grouping is reducing the complexity of cross-channel correlations, by modeling it separately for each group.
Clearly, as each group contains a random subset of channels, not all possible correlations are accounted for.
This is mitigated by channel shuffling and channel concatenation, which makes sure that the channels are grouped altogether albeit in a different order.
As such, the next Timeception layer will group a different subset of channels.
Together, channel grouping and channel shuffling is more cost-effective operation to learn cross-channel correlations than $1\timesnarrow1$ 2D convolutions~\cite{chollet2016xception}.

\begin{figure}[!ht]
\begin{center}
\includegraphics[trim=0mm 15mm 0mm 5mm,width=0.9\linewidth]{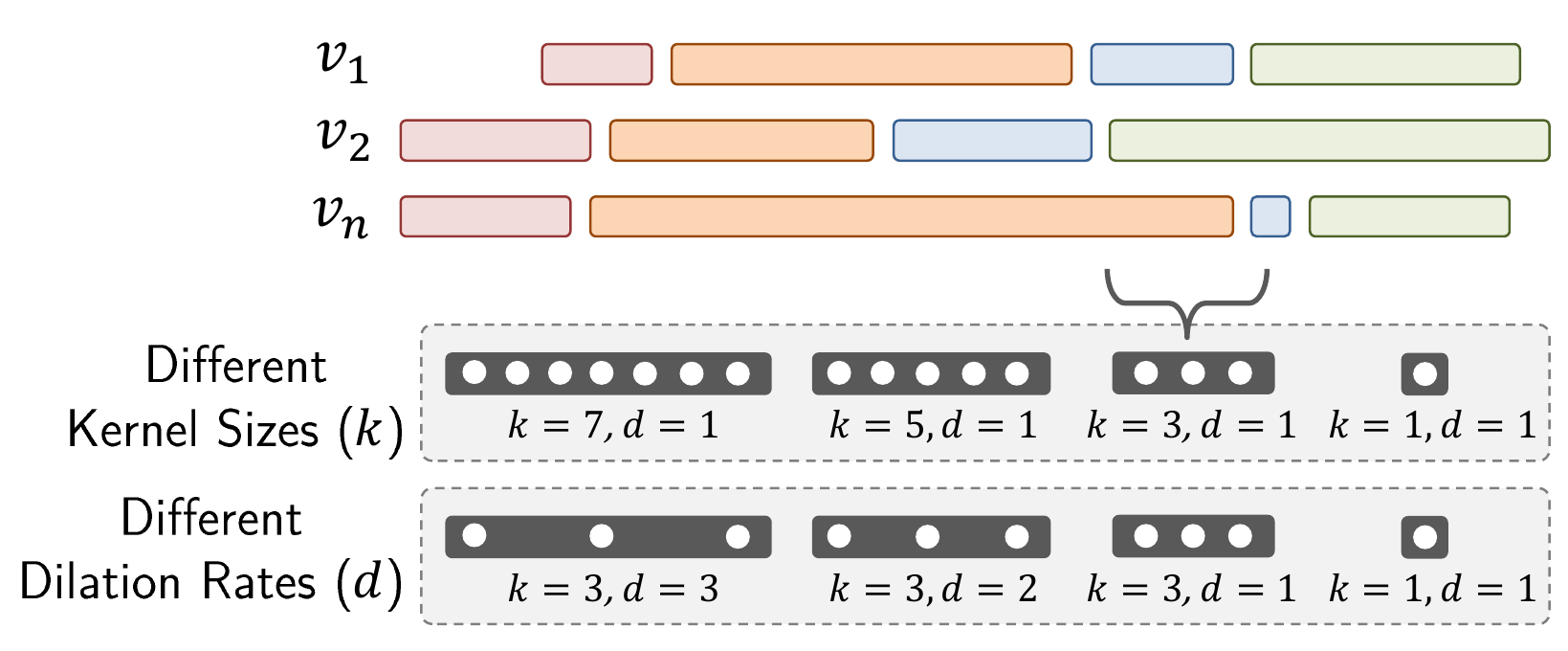}
\end{center}
\caption{To tolerate temporal extents, we use multi-scale temporal kernels, with two options:
\textit{i}. different kernel sizes $k \in \{1,3,5,7\}$ and fixed dilation rate $d=1$,
\textit{ii}. different dilation rates $d \in \{1,2,3\}$ and fixed kernel size $k=3$.}
\label{fig:3-4}
\vspace*{-5mm}
\end{figure}

\partitle{Tolerating Variant Temporal Extents.}
The second objective for the Timeception layer is to tolerate the differences in temporal extents of complex actions.
While in the previous description we assume a fixed length for the \emph{temporal-only} kernels, one-actions in a complex video may vary in length.
To this end, we propose to replace fixed-size temporal kernels with multi-scale temporal kernels.
There are two possible ways to implement multi-scale kernels, see figure~\ref{fig:3-4}.
The first way, inspired by Inception~\cite{szegedy2015going} for images, is to adopt $K$ kernels, each of a different size $k$.
The second way, inspired by~\cite{van2016wavenet}, is to employ dilated convolutions.

The temporal convolution module, see figure~\ref{fig:3-1}{\color{red}b}, takes as an input the features of one group $\mathbf{X}_n \in \mathbb{R}^{T \timesnarrow L \timesnarrow L \timesnarrow [C/N]}$. Then it applies five temporal operations in total.
The first three operations are temporal convolutions with kernel sizes $k=\{3, 5, 7\}$, each maintaining the number of channels at $C/N$.
The forth operation is a temporal max-pooling with stride $s=1$ and kernel size $k=2$.
Its purpose is to max-out activations over local temporal window $(k=2)$, instead of convolving them.
The fifth operation is simply a dimension reduction for the input feature $\mathbf{X}_n$, using a $1 \timesnarrow 1$ spatial convolution.
To maintain a manageable number of dimensions for the output, the input to the first fours operations are shrinked by a factor of $M$ using a $1 \timesnarrow 1$ spatial convolution .
After the channel reduction, all five outputs are concatenated across channel dimension, resulting in an output $\mathbf{Y}_n \in \mathbb{R}^{T \times L \times L \times (5C/MN)}$.

\partitle{Summary of Timeception.}
A Timeception layer, see figure~\ref{fig:3-1}{\color{red}a}, expects an input feature $\mathbf{X} \in \mathbb{R}^{T \times L \times L \times C}$ from the previous layer in the network.
The features $\mathbf{X}$ across the channel dimension are then split into $N$ channel groups.
Each group $\mathbf{X}_n \in \mathbb{R}^{T \times L \times \times L \times [C/N]}$ is convolved with the \emph{temporal convolution module}, resulting in $\mathbf{Y}_n \in \mathbb{R}^{T \times L \times \times L \times [5C/MN]}$.
This module expands the number of channels per group by a factor of $5/M$.
After that, the features of all groups $\mathbf{Y} = \{ \mathbf{Y}_n \mid n \in [1, ..., N]\}$ are concatenated across the channel axis and then randomly shuffled.
Last, to adhere to the second design principle of \emph{subspace balance}, the Timeception layer concludes with a temporal max pooling of kernel size $k=2$ and stride $s=2$.
The reason is that while the channel subspace expands by a factor of $5/M$ after each Timeception layer, the temporal subspace shrinks by a factor of $2$.

\subsection{The Final Model}
The final model consists of four Timeception layers stacked on top of the last convolution layer of a CNN, used as backbone.
We explore two backbone choices: a spatial 2D CNN and a short-range spatiotemporal 3D CNN.

\partitle{2D CNN.}
The first baseline uses ResNet-152~\cite{he2016deep} as backbone.
It takes as an input $128$ video frames, and processes them, up to the last spatial convolution layer \texttt{res5c}.
Thus, the corresponding output for the input frames is the feature $\mathbf{X} \in \mathbb{R}^{128 \times 7 \times 7 \times 2048}$.
Then, we proceed with four successive layers of Timeception, with BatchNorm and ReLU.
Each has channel expansion factor of $5/M = 5/4 = 1.25, M=4$ and temporal reduction factor of $2$.
Thus, the resulting feature is $\mathbf{Y} \in \mathbb{R}^{8 \times 7 \times 7 \times 5000}$.
To further reduce the spatial dimension, we follow the convention of CNNs by using spatial average pooling, which results in the feature $\mathbf{Y^{\prime}} \in \mathbb{R}^{8 \times 5000}$.
And to finally reduce the temporal dimension, we use depthwise-separable temporal convolution with kernel size $k \in \mathbb{R}^{8\timesnarrow 1 \timesnarrow 1 \timesnarrow 1}$ with no zero-padding.
The resulted feature $\mathbf{Z} \in \mathbb{R}^{5000}$ is classified with a two-layer MLP, with BatchNorm and ReLU.

\partitle{3D CNN.}
The second baseline uses I3D~\cite{carreira2017quo} as backbone.
It takes as an input $128$ video segments (each has 8 successive frames), and independently processes these segments, up to the last spatiotemporal convolution layer \texttt{mixed-5c}.
Thus, the corresponding output for the input segments is the feature $\mathbf{X} \in \mathbb{R}^{128 \times 7 \times 7 \times 1024}$.
The rest of this baseline is no different than the previous one.
The benefit of using I3D is that the Timeception layers learn long-range temporal combinations of short-range spatiotemporal patterns.

\partitle{Implementation.}
When training the model on a specific dataset, first we pretrain the backbone CNN on this dataset.
We use uniformly sampled frames for the 2D backbone and uniformly sampled video segments (each has 8 successive frames) for the 3D backbone.
After pre-training, we plug-in Timeception and MLP layers on top of the last convolution layer of the backbone and fine-tune the model on the same dataset.
At this stage, only Timeception layers are trained, while the backbone CNN is frozen.
The model is trained with batch-size 32 for 100 epoch.
It is optimized with SGD with $0.1$, $0.9$ and $1e$-$5$ as learning rate, momentum and weight decay, respectively.
Our public implementation~\cite{timeceptioncode} uses TensorFlow~\cite{tensorflow2015-whitepaper} and Keras~\cite{chollet2015keras}.

\section{Experiments}\label{sec:experiments}
\subsection{Datasets}
The scope of this paper is complex actions with their three properties: composition, temporal extent and temporal order --see figure~\ref{fig:1-1}. Thus, we choose to conduct our experiments on Charades~\cite{sigurdsson2016hollywood}, Breakfast Actions~\cite{kuehne2014language} and MultiTHUMOS~\cite{yeung2018every}.
Other infamous datasets for action recognition do not meet the properties of complex actions.

\partitle{Charades}
is multi-label, action classification, video dataset with 157 classes.
It contains 8k, 1.2k and 2k videos for training, validation and test splits, respectively (67 hrs for training split).
On average, each complex action (\textit{i.e.} each video) is 30 seconds and contains 6 one-actions.
Thus, Charades meets the criteria of complex actions.
We use mean Average Precision (mAP) for evaluation.
As labels of test set are held out, we report results on the validation set, similar to all related works~\cite{sigurdsson2017asynchronous, girdhar2017actionvlad,sigurdsson2017asynchronous,wang2017non,wang2018videos}.

\partitle{Breakfast Actions}
is a dataset for unscripted cooking-oriented human activities.
It contains 1712 videos in total, 1357 for training and 335 for test.
The average length of videos is 2.3 minutes.
It is a video classification task of 12 categories of breakfast activities, where each video represents only one activity.
Besides, each video has temporal annotation of one-actions composing its activity.
In total, there are 48 classes of one-actions.
In our experiments, we only use the activity annotation, and we do not use the temporal annotation of the one-actions.

\partitle{MultiTHUMOS}
is a dataset for human activities in untrimmed videos, with the primary focus on temporal localization.
It contains 65 action classes and 400 videos (30 hrs).
Each video can be thought of a complex action, which comprises 11 one-actions on average.
MultiTHUMOS extends the original THUMOS-14~\cite{idrees2017thumos} by providing multi-label annotation for the videos in validation and test splits.
Having multiple and dense labels for the video frames enable temporal models to benefit from the temporal relations between one-actions across the video.
Similar to Charades, mAP is used for evaluation.

\subsection{Tolerating Temporal Extents}
In this experiment, we evaluate the capacity of the multi-scale kernels to tolerate the differences in temporal extents of actions.
The experiment is carried out on Charades.

\begin{figure}[!ht]
\begin{center}
\includegraphics[trim=2mm 8mm 2mm 2mm,width=1.0\linewidth]{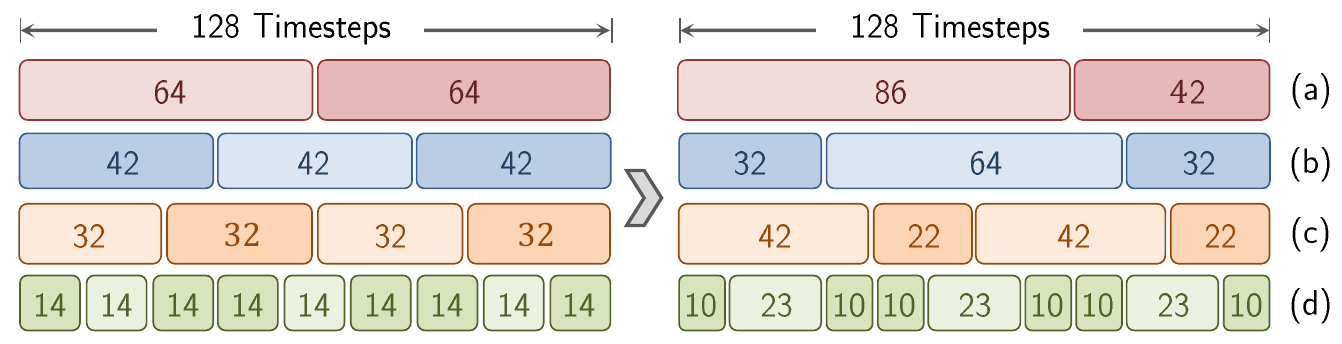}
\end{center}
\caption{
We split a video of 128 timesteps into segments of equal length (left, before alteration), and alter their temporal extents by expansion and shrinking (right, after alteration).
We use 4 types of alterations:(a) very-coarse, (b) coarse, (c) fine, and (d) very-fine.
Numbers in boxes are timesteps.}
\label{fig:4-1}
\vspace*{-0mm}
\end{figure}

\partitle{Original v.s. Altered Temporal Extents}
First, we train two baselines, one with multi-scale temporal kernels (as in Timeception) and the other with fixed-size kernels.
The training is done on the original temporal extent of training videos.
Then, \emph{at test time only}, we alter the temporal extents of test videos.
Specifically, we split each test video into segments.
Then, we temporally expand or shrink these segments.
Expansion is done by repeating frames, while shrinking is done by dropping frames.
We use 4 types of alterations with varying granularity to test the model in different scenarios: (a) very-coarse, (b) coarse, (c) fine, and (d) very-fine, see in figure~\ref{fig:4-1}.

\begin{table}[!ht]
\centering
\renewcommand{\arraystretch}{1.0}
\setlength\tabcolsep{5.4pt}
\begin{tabular}{lcccc}
\specialrule{0.3mm}{.0em}{.3em}
Altered Extent & \multicolumn{4}{c}{Percentage Drop $\downarrow$ in mAP} \\
\cmidrule(lr){1-1} \cmidrule(lr){2-5}
& \multicolumn{2}{c}{I3D} & \multicolumn{2}{c}{ResNet} \\
& Fixed $\downarrow$ & Multi  $\downarrow$ & Fixed  $\downarrow$ & Multi  $\downarrow$ \\
\cmidrule(lr){2-3}  \cmidrule(lr){4-5}
(a) very-coarse    & 2.09 & 1.75 & 1.52 & 1.08 \\
(b) coarse         & 2.92 & 2.44 & 3.26 & 2.15 \\
(c) fine           & 1.74 & 1.12 & 1.59 & 1.13 \\
(d) very-fine      & 2.18 & 1.71 & 1.38 & 1.20 \\
\specialrule{0.3mm}{.0em}{.0em}
\end{tabular}
\caption{
Timeception, with multi-scale kernel, tolerates the altered temporal extents better than fixed-size kernels.
We report the percentage drop in mAP (lower is better) when testing on original \textit{v.s.} altered videos of Charades.
I3D and ResNet are backbone CNNs.}
\label{tbl:4-1}
\vspace*{-5pt}
\end{table}

The results of this controlled experiment are shown in table~\ref{tbl:4-1}. We observe that Timeception is more effective than fixed-size kernels in handling unexpected variations in temporal extents.
The same observations is confirmed using either I3D or ResNet as backbone architecture.

\partitle{Fixed-size vs. Multi-scale Temporal Kernels}
This experiment points out the merit of using multi-scale temporal kernels. For this, we compare fixed-size temporal convolutions against multi-scale temporal-only convolutions, either with different kernel sizes $k$ or dilation rates $d$.
And we train 3 baseline models with different configurations of $k, d$:
\textit{i}. Fixed kernel size and fixed dilation rate $d=1, k=3$.
This is the typical configuration used in 3D CNNs~\cite{tran2015learning,carreira2017quo,wang2017non,xie2017rethinking}.
\textit{ii}. Different kernel sizes $k \in \{1, 3, 5, 7\}$ and fixed dilation rate $d=1$.
\textit{iii}. Fixed kernel size $k = 3$ and different dilation rates $d \in \{1, 2, 3\}$.

The result of this experiment are shown in table~\ref{tbl:4-2}.
We observe that using multi-scale kernels is better suited for modeling complex actions than fixed-size kernels.
The same observation holds for both I3D and ResNet as backbones.
Also, we observe little to no change in performance when using different dilation rates $d$ instead of different kernel sizes $k$.

\begin{table}[!ht]
\centering
\renewcommand{\arraystretch}{1.0}
\setlength\tabcolsep{7pt}
\begin{tabular}{ccccc}
\specialrule{0.3mm}{.0em}{.3em}
Kernel      & Kernel         & Dilation      & \multicolumn{2}{c}{mAP (\%)} \\
\cmidrule(lr){4-5}
Type        & Size ($k$)     & Rate ($d$)    &  ResNet & I3D\\
\midrule
\multirow{2}{*}{Multi-scale} & 1,3,5,7  & 1  & 30.82   & 33.76 \\
            & 3              & 1,2,3         & 30.37   & 33.89 \\
\midrule
Fixed-size  & 3              & 1             & 29.30   & 31.87 \\
\specialrule{0.3mm}{.0em}{.0em}
\end{tabular}
\caption{
Timeception, using multi-scale kernels (i.e. different kernel sizes ($k$) or dilation rates ($d$), outperforms fixed-size kernels on Charades.
I3D/ResNet are backbone.}
\label{tbl:4-2}
\vspace*{-15pt}
\end{table}

\begin{figure*}[!ht]
\begin{center}
\includegraphics[trim=5mm 10mm 2mm 5mm,width=1.0\linewidth]{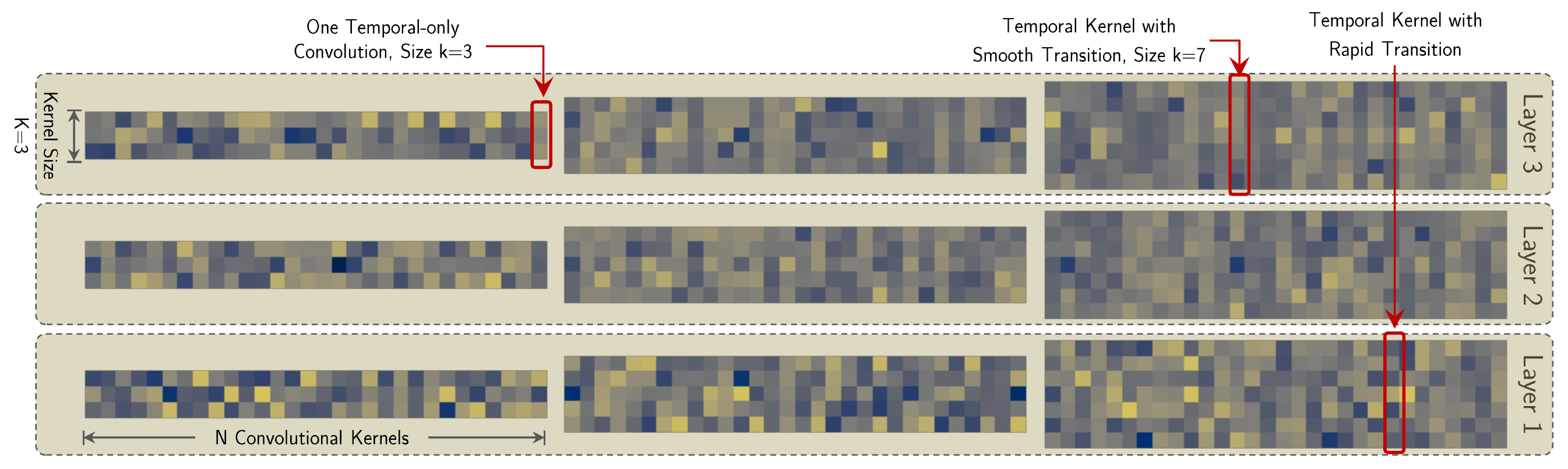}
\end{center}
\caption{The learned weights by temporal convolutions of three Timeception layers. Each uses multi-scale convolutions with varying kernel sizes $k \in \{3, 5, 7\}$. In bottom layer (1), we notice that long kernels ($k=7$) captures fine-grained temporal dependencies. But at the top layer (3), the long kernels tend to focus on coarse-grained temporal correlation. The same behavior prevails for the shot ($k=3$) and medium ($k=5$) kernels.}
\label{fig:4-5}
\vspace*{-5mm}
\end{figure*}

\subsection{Long-range Temporal Dependencies}
In this experiment, we demonstrate the capacity of multiple Timeception layers to learn long-range temporal dependencies for complex actions.
We train several baseline models equipped with Timeception layers. These baselines use different number of input timesteps.
We experiment on Charades, with both ResNet and I3D as backbone.

\partitle{ResNet} is used, with a different number of timesteps as inputs: $T \in \{32, 64, 128\}$, followed by Timeception layers.
ResNet processes one frame at a time.
Hence, in one feedforward pass, the number of timesteps consumed by Timeception layers is equal to that consumed by ResNet.

\partitle{I3D} is considered, with a different number of timesteps as inputs: $T \in \{256, 512, 1024\}$, followed by Timeception layers.
I3D processes $8$ frames into one \textit{super-frame} at a time.
Thus, Timeception layers model $T^{\prime} \in \{32, 64, 128\}$ super-frames, Practically however, as each super-frame is related to a segment of $8$ frames, both I3D+Timeception process in total $T \in \{256, 512, 1024\}$ frames.

\begin{table}[!ht]
\centering
\renewcommand{\arraystretch}{1.0}
\setlength\tabcolsep{2.3pt}
\begin{tabular}{cccccc}
\specialrule{0.3mm}{.0em}{.3em}
\multicolumn{2}{c}{Baseline} & CNN Steps & TC Steps & Params & mAP (\%) \\
\midrule
       & + 3 TC & 32  & 32   & 3.82  & 30.37  \\
ResNet & + 3 TC & 64  & 64   & 3.82  & 31.25  \\
       & + 4 TC & 128 & 128  & 5.58  & 31.82  \\
\midrule
    & + 3 TC & 256  & 32     & 1.95  & 33.89  \\
I3D & + 3 TC & 512  & 64     & 1.95  & 35.46  \\
    & + 4 TC & 1024 & 128    & 2.83  & 37.19  \\
\specialrule{0.3mm}{.0em}{.0em}
\end{tabular}
\caption{
Timeception layers allow for deep and efficient temporal models, able to learn the temporal abstractions needed to learn complex actions.
Columns are:
\textit{Baseline}: backbone CNN + how many Timception layers (TC) on top of it,
\textit{CNN Steps}: input timesteps to the CNN,
\textit{TC Steps}: input timesteps to the first Timeception layer,
\textit{Params}: number of parameters used by Timeception layers, in millions.}
\label{tbl:4-3}
\vspace*{-5pt}
\end{table}

We report results in table~\ref{tbl:4-3} and we make two observations.
First, stacking Timeception layers leads to an improved accuracy when using both ResNet and I3D as backbone.
As the only change between these models is the number of Timeception layers, we deduce that the Timeception layers have succeeded in learning temporal abstractions.
Second, despite stacking more and more Timeception layers, the number of parameters is controlled. 
Interestingly, using 4 Timeception layers on I3D  processing 1024 timesteps requires half the parameters needed for a ResNet processing 128 timesteps.
The reason is the number of channels from ResNet is twice as much as from I3D (2048 \textit{v.s.} 1024).
We conclude that Timeception layers allow for deep and efficient models, able to learn long-range temporal abstractions, which is crucial for complex actions.

\ptspace
\partitle{Learned Weights of Timeception.}
Figure~\ref{fig:4-5} visualizes the learned weights by our model.
Specifically, three Timeception layers trained on top of I3D backbone.
The figure depicts the weights of multi-scale temporal convolutions with different kernel sizes $k \in \{3, 5, 7\}$.
For simplicity, only the first 30 kernels from each kernel-size, are shown.
We make two remarks for these learned weights.
First, at layer 1, we notice that long kernels ($k=7$) captures fine-grained temporal dependencies, because of the rapid transition of kernel weights.
But at layer 3, these long kernels tend to focus on coarse-grained temporal correlations, because of the smooth transition between kernel weights.
The same behavior prevails for the short ($k=3$) and medium ($k=5$) kernels.
Second, at layer 3, we observe that long-range and short-range temporal patterns are learned by short kernels ($k=3$) and long kernels ($k=7$), respectively.
The conclusion is that for complex actions, both video-wide and local temporal reasoning, even at the top layer, is crucial for recognition.

\subsection{Effectiveness of Timeception}
\begin{figure}[ht]
\begin{center}
\includegraphics[trim=0mm 8mm 0mm 0mm,width=0.8\linewidth]{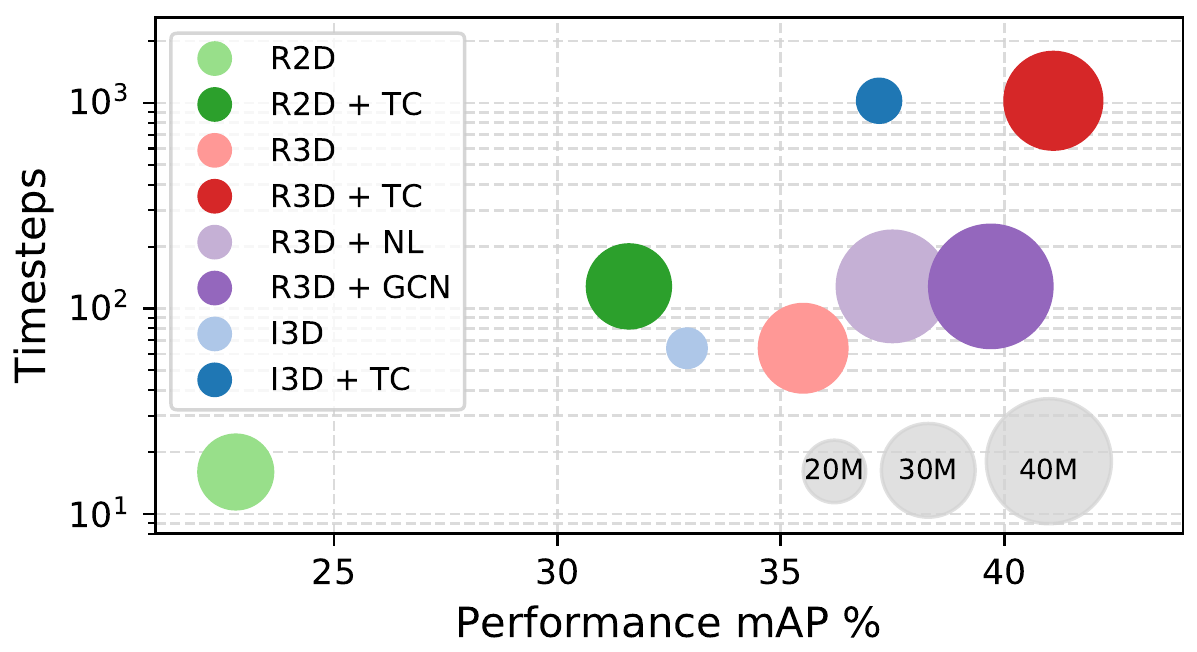}
\end{center}
\caption{
Top: the cost of adding new Timeception layers is marginal, compared to related temporal layers.
Bottom: Timeception improves performance, scales up temporal capacity of backbone CNNs while maintaining the model size.}
\label{fig:4-9}
\vspace*{-5mm}
\end{figure}

To demonstrate the effectiveness of Timeception, we compare it against related temporal convolution layers:
\textit{i}. separable temporal convolution~\cite{tran2018closer}, that models both $\mathcal{T}, \mathcal{C}$ simultaneously.
\textit{ii}. grouped separable temporal convolution to model $\mathcal{T}$, followed by $1 \timesnarrow 1$ 2D convolution to model $\mathcal{C}$.
\textit{iii}. grouped separable temporal convolution to model $\mathcal{T}$, followed by channel shuffling to model $\mathcal{C}$.
Interestingly in figure~\ref{fig:4-9} top, Timeception is very efficient in maintaining a reasonable increase in number of parameters as the network goes deeper.
Also, figure~\ref{fig:4-9} bottom shows how Timeception improves mAP on Charades, scales up temporal capacity of backbone CNNs while maintaining the overall model size.

\begin{figure}[ht]
\begin{center}
\includegraphics[trim=-2mm 0mm 0mm 0mm,width=0.8\linewidth]{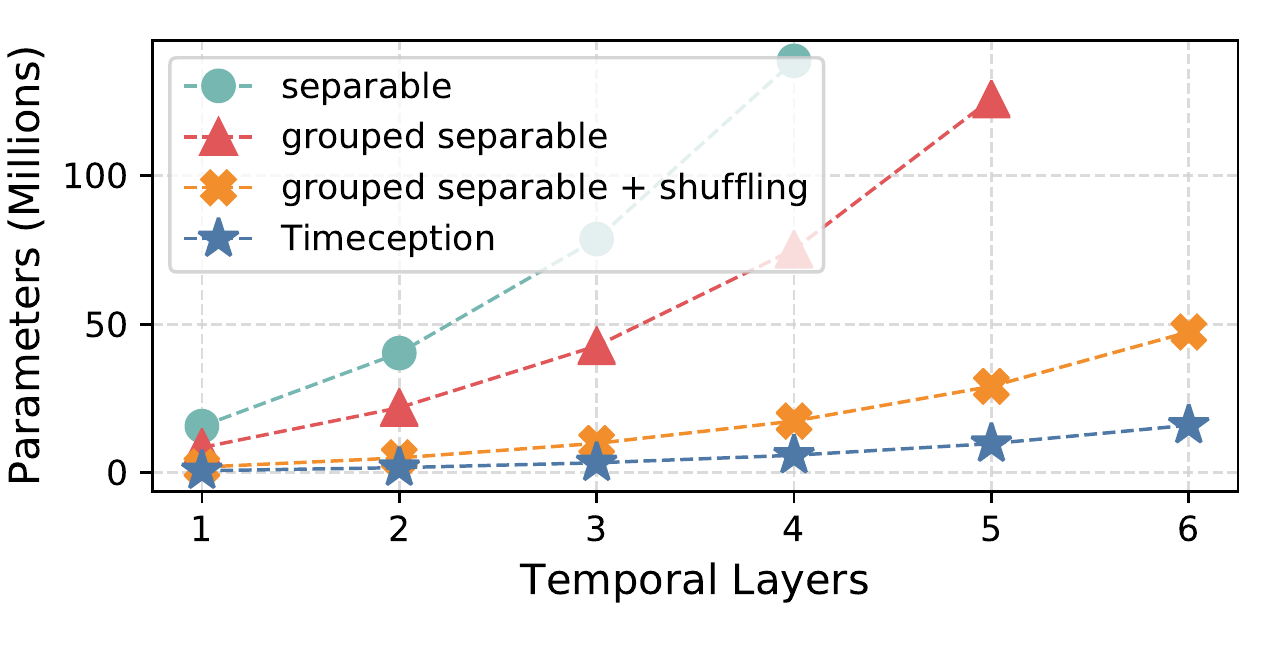}
\end{center}
\vspace*{-10mm}
\end{figure}

\subsection{Experiments on Benchmarks}

\partitle{Charades}
is used to evaluate our model, and to compare against related works.
In this experiment, our baseline networks use 4 Timeception layers.
The number of convolutional groups is 8 for I3D and 16 for ResNet, be it 2D or 3D.
The results in table~\ref{tbl:4-6} shows that Timeception monotonically improves the performance of the backbone CNN.
The absolute gain on top of ResNet and I3D is $8.8\%$ and $4.3\%$, respectively.

\begin{table}[!ht]
\centering
\renewcommand{\arraystretch}{1.0}
\setlength\tabcolsep{1pt}
\begin{tabular}{clcc}
\specialrule{0.3mm}{.0em}{.3em}
Ours    & $\;\;\;\;\;\;\;$Method           						& Modality 		& mAP (\%) \\
\midrule
		&\cite{sigurdsson2017asynchronous} Two-stream			& RGB + Flow 	& 18.6 \\
		&\cite{sigurdsson2017asynchronous} Two-stream + LSTM	& RGB + Flow 	& 17.8 \\
		&\cite{girdhar2017actionvlad} ActionVLAD				& RGB + iDT		& 21.0 \\
		&\cite{sigurdsson2017asynchronous} Temporal Fields		& RGB + Flow   	& 22.4 \\
		&\cite{zhou2017temporal} Temporal Relations				& RGB		   	& 25.2 \\
\midrule
		&\cite{charades2017algorithms} ResNet-152				& RGB        	& 22.8 \\
\cmark 	&$\;\;\;\;\;\;\;$ResNet-152 + TC						& RGB        	& 31.6 \\
\midrule
		&\cite{carreira2017quo} I3D								& RGB        	& 32.9 \\
\cmark	&$\;\;\;\;\;\;\:$I3D + TC								& RGB        	& 37.2 \\
\midrule
		&\cite{wang2017non} 3D ResNet-101						& RGB        	& 35.5 \\
		&\cite{wang2017non} 3D ResNet-101 + NL					& RGB        	& 37.5 \\
		&\cite{wang2018videos} 3D ResNet-50 + GCN			    & RGB + RP    	& 37.5 \\
		&\cite{wang2018videos} 3D ResNet-101 + GCN		        & RGB + RP	    & 39.7 \\
\cmark	&$\;\;\;\;\;\;\:$3D ResNet-101 + TC						& RGB        	& 41.1 \\
\specialrule{0.3mm}{.0em}{.0em}
\end{tabular}
\caption{
Timeception (TC) outperforms related works using the same backbone CNN. It achieves the absolute gain of $8.8\%$ and $4.3\%$ over ResNet and I3D, respectively. More over, using the full capacity of Timeception improves $1.4\%$ over best related work.}
\label{tbl:4-6}
\vspace*{-10pt}
\end{table}

\begin{figure}[ht]
\begin{center}
\includegraphics[trim=-8mm 14mm -6mm 5mm,width=1.0\linewidth]{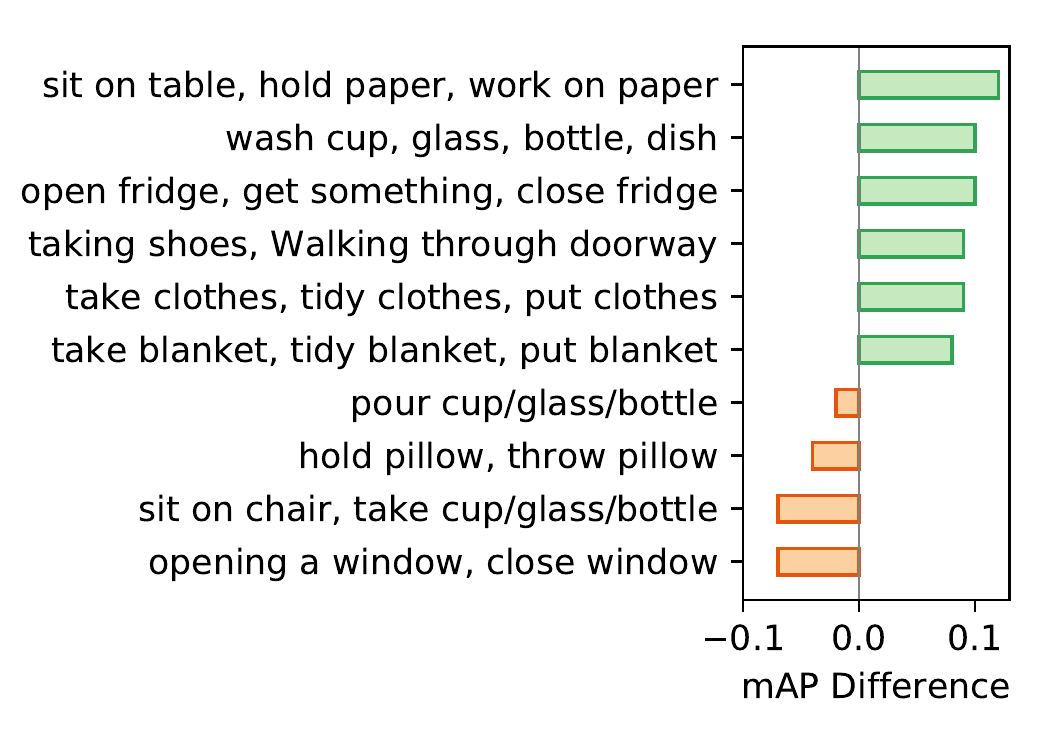}
\end{center}
\caption{
Multi-scale Timeception outperforms the fixed-kernel when complex actions are dynamic, in green.
But when complex actions with rigid temporal patters, fixed-size performs better than multi-scale, in orange.}
\label{fig:4-7}
\vspace*{-3mm}
\end{figure}

\begin{figure}[ht]
\begin{center}
\includegraphics[trim=-8mm 14mm -6mm 5mm,width=1.0\linewidth]{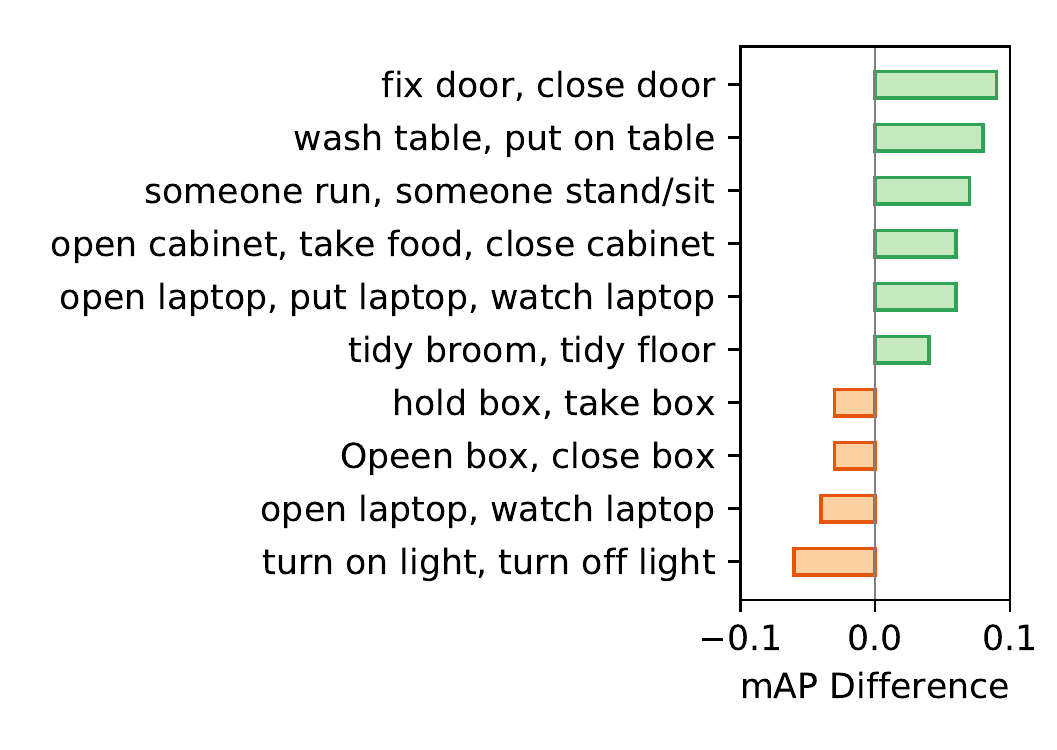}
\end{center}
\caption{
Long-range Timeception outperforms the short-range version when complex actions need the entire video to unfold, in green.
However, we see one-actions that short-range Timeception can easily capture, in orange.}
\label{fig:4-8}
\vspace*{-3mm}
\end{figure}

Beyond the overall mAP, how beneficial is Timeception? And in what cases exactly does it help?
To answer this question, we make two comparisons to assess the relative performance of Timeception.
We experiment two scenarios:
\textit{i}. short-range ($32$ timesteps) v.s. long-range ($128$ timesteps),
\textit{ii}. fixed-scale v.s. multi-scale kernels.
The results are shown in figures~\ref{fig:4-7},~\ref{fig:4-8}, and we make two observations.

First, when comparing the relative performance of multi-scale vs. fixed-size Timeception, see figure~\ref{fig:4-7}, we observe that multi-scale Timeception excels in complex actions with dynamic temporal patterns.
As an example, ``take clothes + tidy clothes + put clothes", one actor may take longer than others to tidy clothes.
In contrast, fixed-size Timeception excels in the cases where the complex action is more rigorous in the temporal pattern, \emph{e.g.} ``open window + close window".
Second, when comparing the relative performance of short-range ($32$ timesteps) v.s. long-range ($1024$ timesteps) Timeception, see figure~\ref{fig:4-8}, the later excels in complex actions than requires the entire video to unfold, \textit{e.g.} ``fix door + close door".
However, short-range Timeception would do better in one-actions, like ``open box + close box" or ``turn on light + turn of light".

\partitle{Breakfast Actions}
is used as a second dataset to experiment our model.
The average length of a video in this datataset is 2.3 sec.
For this experiment, we use 3 layers of Timeception.
And as for the backbone, we use I3D and 3D ResNet-50.
None of the backbones is fine-tuned on this dataset, only Timeception layers are trained.
To make one video consumable by our baseline, from each video we uniformly sample 64 video snippet, each of 8 sucessive frames.
That makes the total timesteps modeled by the baseline is 512.
Finally, we report result in table~\ref{tbl:4-7}.

\begin{table}[!ht]
\centering
\renewcommand{\arraystretch}{1.0}
\setlength\tabcolsep{1pt}
\begin{tabular}{lcc}
\specialrule{0.3mm}{.0em}{.3em}
Method & Activities (Acc. \%)  & Actions (mAP \%) \\
\midrule
I3D						    & 64.31 & 47.71 \\
I3D + TC        		    & 69.30 & 56.36 \\
\midrule
3D ResNet-50		        & 66.73 & 53.27  \\
3D ResNet-50 + TC		    & 71.25 & 59.64 \\
\specialrule{0.3mm}{.0em}{.0em}
\end{tabular}
\caption{
Timeception outperform baselines in recognizing the long-range activities of Breakfast dataset.}
\label{tbl:4-7}
\vspace*{-10pt}
\end{table}

\partitle{MultiTHUMOS}
is used as a third dataset to experiment our model.
This helps in investigating the generality on different datasets.
Related works use this dataset for temporal localization of one-actions in each video of complex action. Differently, we use this dataset to serve our objective: multi-label classification of complex actions, \textit{i.e.} the entire video.
As such, the evaluation method used is mAP~\cite{scikit-learn}.
To assess the performance of our model, we compare against I3D as a baseline. As shown in the results in table~\ref{tbl:4-8}, Timeception, equipped with multi-scale kernels outperforms that with fixed-size kernel.

\begin{table}[!ht]
\centering
\renewcommand{\arraystretch}{1.0}
\setlength\tabcolsep{8pt}
\begin{tabular}{lccc}
\specialrule{0.3mm}{.0em}{.3em}
\multirow{2}{*}{Method} & Kernel 	& Dilation 	& \multirow{2}{*}{mAP (\%)} \\
						& Size $k$	& Rate $d$	& \\
\midrule
I3D						& 	--		& 	 --		& 72.43 \\
I3D + Timeception		& 3 		& 	1		& 72.83 \\
I3D + Timeception		& 3 		& 	1,2,3	& 74.52 \\
I3D + Timeception		& 1,3,5,7 	& 	1		& 74.79 \\
\specialrule{0.3mm}{.0em}{.0em}
\end{tabular}
\caption{
Timeception, with multi-scale temporal kernels, helps baseline models to capture the long-range dependencies between one-actions in videos of MultiTHUMOS.}
\label{tbl:4-8}
\vspace*{-10pt}
\end{table}

\section{Conclusion}\label{sec:conclusions}
Complex actions such as ``cooking a meal'' or ``cleaning the house'' can only be recognized when processed fully. This is in contrast to one-actions, that can be recognized from a small burst of frames.
This paper presents Timeception, a novel temporal convolution layer for complex action recognition. Thanks to using efficient temporal-only convolutions, Timeception can scale up to minute-long temporal modeling.
In addition, thanks to multi-scale temporal convolutions, Timeception can tolerate the changes in temporal extents of complex actions.
Interestingly, when visualizing the temporal weights we observe that earlier timeception layers learn fast temporal changes, whereas later timeception layers focus on more global temporal transitions.
Evaluating on popular benchmarks, the proposed Timeception improves the state-of-the-art notably.

\subsection*{Acknowledgment}\label{acknowledgement}
We thank Xiaolong Wang~\cite{wang2017non} for sharing the code.

{\small
\bibliographystyle{unsrt}
\bibliography{main}
}

\end{document}